\documentclass[runningheads]{llncs}
\usepackage{graphicx}
\usepackage{amsmath,amssymb,bm} 
\usepackage{color}
\usepackage{enumitem}
\usepackage{caption}
\usepackage[ruled,vlined]{algorithm2e}
\usepackage[width=122mm,left=12mm,paperwidth=146mm,height=193mm,top=12mm,paperheight=217mm]{geometry}

\def\eg{\textit{e.g. }}
\def\etal{\textit{et al. }}

\DeclareMathOperator*{\argmin}{\arg\!\min\,}

\def\I{\mathbf{I}}

\def\y{\mathbf{y}}

\def\x{\mathbf{x}}
\def\z{\mathbf{z}}
\def\M{\mathbf{M}}
\def\W{\mathbf{W}}
\def\d{\mathbf{d}}
\def\L{\mathcal{L}}

\begin{document}
\pagestyle{headings}
\mainmatter

\title{Deep Convolutional Compressed Sensing for LiDAR Depth Completion} 

\author{Nathaniel Chodosh, Chaoyang Wang, Simon Lucey}


\institute{Robotics Institute, Carnegie Mellon University\\
  \email{ \{nchodosh,chaoyanw\}@andrew.cmu.edu, slucey@cs.cmu.edu}
}

\maketitle

\begin{abstract}
  In this paper we consider the problem of estimating a dense depth map from a set of sparse LiDAR points. We use techniques from compressed sensing and the recently developed Alternating Direction Neural Networks (ADNNs) to create a deep recurrent auto-encoder for this task. Our architecture internally performs an algorithm for extracting multi-level convolutional sparse codes from the input which are then used to make a prediction. Our results demonstrate that with only two layers and 1800 parameters we are able to out perform all previously published results, including deep networks with orders of magnitude more parameters.
\end{abstract}

\section{Introduction}
\begin{figure}
\centering
  \includegraphics[width=\textwidth]{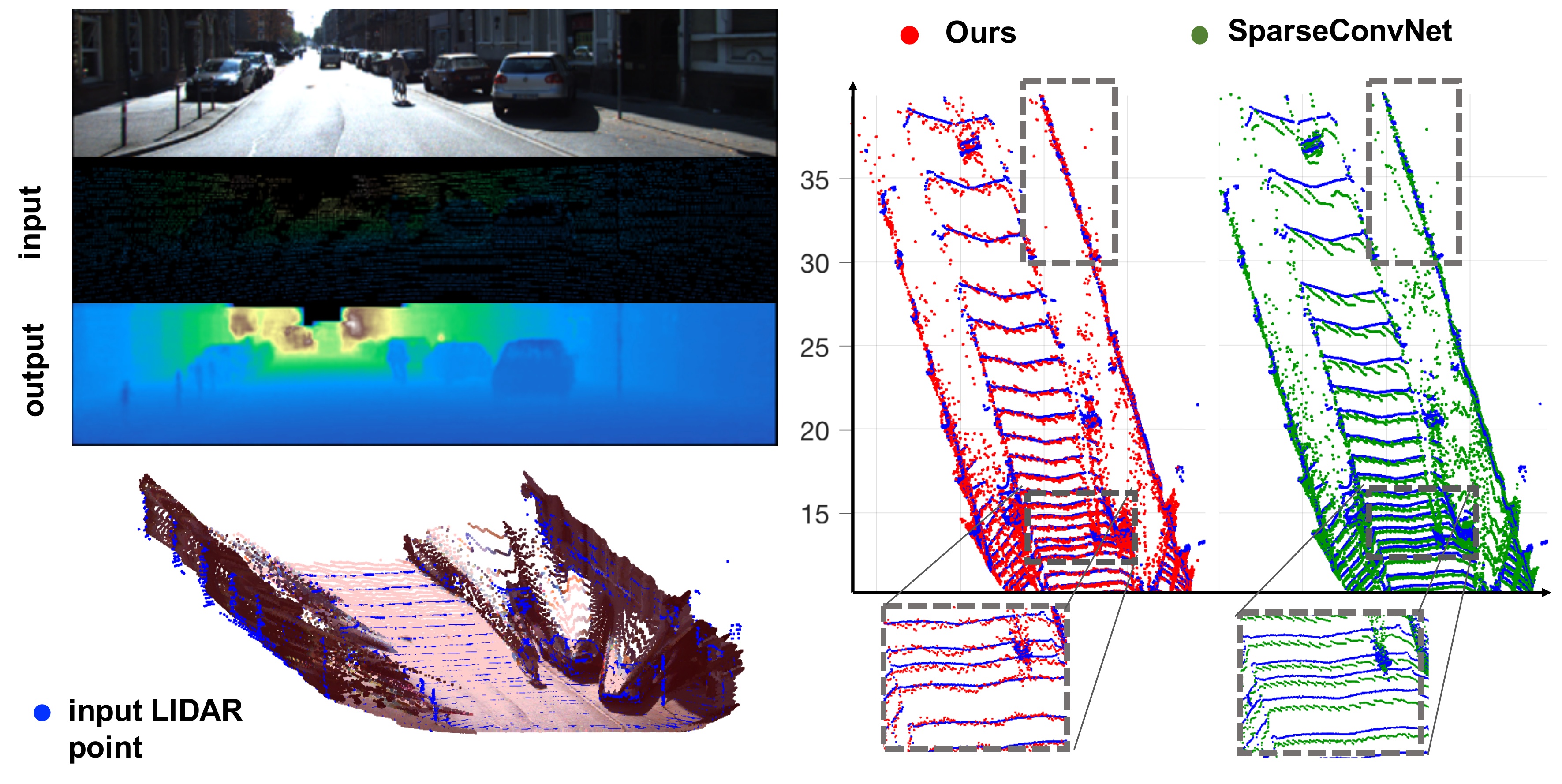}\\
  \caption{The bicyclist and bollards can barely be seen in the input map but a clearly represented in the output. Our method also accurately reconstructs very thin objects such as the sign post. On the right it can be seen that our method enforces that its prediction should match the input points while the SparseConvNet systematically underestimates the depth.}
  \label{fig:intro}
\end{figure}
In recent years 3D information has become an important component of robotic sensing. Usually this information is presented in 2.5D as a depth map, either measured directly using LiDAR or computed using stereo correspondence. Since LiDAR and stereo techniques yield few samples relative to modern image sensors, it has become desirable to convert sparse depth measurements into high resolution depth maps as shown in Figure~\ref{fig:intro}.

Recent works~\cite{sparsetodense,uhrig} have directly applied deep networks to depth completion from sparse measurements. However, common network architectures have two drawbacks when applied to this task: 1) They implicitly pose depth completion as finding a mapping from sparse depth maps to dense ones, instead of as finding a depth map that is consistent with the sparse input. This essentially throws away information and we observe that feed forward networks do not learn to propagate the input points through to the output. Qualitative evidence of this can be seen in Figure (\ref{fig:intro}). 2) Common networks are sensitive to the sparsity of the input since they treat all pixels equally, regardless of whether or not they represent samples or missing input. Special CNN networks have been designed to address this problem, but they still do not express the constraints given by the input~\cite{uhrig}. In this paper we address both of these issues with a novel deep recurrent autoencoder architecture, which internally optimizes its depth prediction with respect to both sparsity and input constraints.
To do this, we have taken inspiration from Compressed sensing (CS) which provides a natural framework for this problem. Formally, CS is concerned with recovering signals from incomplete measurements by enforcing that signals be sparse when measured in an appropriate basis. This basis takes the form of an overcomplete matrix which maps sparse representations to observed signals.

The choice of dictionary is crucial for recovering the signal efficiently, especially when the dimensionality is high. For high resolution imagery data, such as depth maps, multi-layer convolutional sparse coding (CSC)~\cite{papyan} is effective as it explicitly models local interactions through the convolution operator with tractable computational and model complexity. However, none of the existing multi-layer convolutional sparse coding algorithms are designed for learning from sparse ground truth data. This is reflected by the fact that recent works~\cite{hawe2011dense,liu2015depth} applying CS to depth completion are restricted to using single-level, hand crafted dictionaries. CS has also fallen out of fashion since the existing algorithms have difficult to interpret hyper-parameters, and often do not achieve good performance without careful tuning of these parameters.


Recent developments in the formal analysis of deep learning have shown that convolutional neural networks and convolutional sparse coding are closely related. Specifically it has been shown that CNNs with ReLU activation functions are carrying out a specific form of the layered thresholding algorithm for CSC. Layered thresholding is a simple algorithm for solving multi-layered convolutional sparse coding (ML-CSC) problems, which can be effective when there is little noise and the coherence of the dictionary is high. Motivated by the work of Murdock \etal~\cite{murdock}, in this paper we propose a network architecture which encodes a more sophisticated algorithm for ML-CSC. Encoding the ML-CSC objective in a deep network allows us to learn the dictionaries and parameters together in an end to end fashion. We show that by better approximating this objective, we can out perform all published results on the KITTI depth completion benchmark while using far fewer parameters and layers. Furthermore, this work builds on the Alternating Direction Neural Network (ADNN) framework of Murdock \etal which gives theoretical insight into deep learning and we believe is a promising new area of research.

To summarize, the main contributions of this paper are:
\begin{enumerate}
\item We frame an end-to-end multi-layer dictionary learning algorithm as a neural network. This allows us to effectively learn dictionaries and hyper-parameters from a large dataset. In comparison, existing CS algorithms either use hand crafted dictionaries, separately learned multi-level dictionaries, or are inapplicable to incomplete training data~\cite{sulam}, as is our case.

\item Our method allows for explicit encoding of the constraints from the input sparse depth. Current deep learning approaches~\cite{sparsetodense} simply feed in a sparse depth map and rely solely on data to teach the network to identify which inputs represent missing data. Some recent models~\cite{uhrig} explicitly include masks to achieve sparsity invariance, but none have a guaranteed way of encoding that the input is a noise corrupted subset of the desired output. In contrast our method directly optimizes the predicted map with respect to the input.

\item Our method demonstrates state-of-the-art performance with much fewer parameters compared to deep networks. In fact, using only two layers of dictionaries and 1600 parameters, our method already substantially outperforms modern deep networks which use more than 20 layers and over 3 million parameters~\cite{sparsetodense}. As a result of having fewer parameters, our approach trains faster and requires less data.
  
\end{enumerate}
  
\section{Related Work}
\label{sec:related-work}

Since our proposed method is a fusion of deep learning and compressed sensing, we will review in this section previous work that has used either technique for depth estimation.

\subsection{Compressed Sensing}
\label{sec:compressed-sensing}


Compressed sensing is a technique in signal processing for recovering signals from a small set of measurements. Naturally it has been applied to depth completion in previous work, but has been limited to single-level hand-crafted dictionaries. The earliest is Hawe \etal\cite{hawe2011dense}, who show that disparity maps can be represented sparsely using the wavelet basis. L.-K. Liu \etal\cite{liu2015depth} built on that by combining wavelets with contourlets and investigated the effect of different sampling patterns. Both methods were out performed by Ma \& Karaman\cite{ma2016sparse} who exploit the simple structures of man-made indoor scenes to achieve full depth reconstruction. In contrast to all of these works, our approach learns multi-level convolutional dictionaries from a large dataset of incomplete ground truth depth maps.

\subsection{Deep Learning}
\label{sec:depth-upsampling}

Depth estimation using deep learning has largely been restricted to single-shot, RGB to depth prediction. This line of inquiry started with Eigen \etal\cite{eigen2014depth} who showed that a deep network could reasonably estimate depth using only an RGB image. Many variants of this method have since been explored\cite{kuznietsov2017semi,godard2016unsupervised,liu2015deep}. Lania \etal~\cite{laina2016deeper} introduced up-projection blocks which allowed for very deep networks and several other works have proposed variants of their architecture. The most relevant of these variants is the Sparse-to-Dense network of Ma \& Karaman\cite{sparsetodense}, which they also apply to depth completion from LiDAR points. Uhrig \etal\cite{uhrig} introduced the KITTI depth completion dataset, and showed that CNNs which explicitly encode the sparsity of the input achieve much better performance. Riegler \etal\cite{riegler} designed ATGV-Net, a deep network for depth map super resolution, but they assume a rectangular grid of inputs so it is not applicable to LiDAR completion. We will use the methods of Ma \& Karaman and Urhig \etal as our baseline comparisons since they represent the state-of-the-art in LiDAR depth completion
\\
\\
\noindent\textbf{Notations.} We define our notations throughout as follows: lowercase boldface symbols (\eg~$\mathbf{x}$) denote vectors, uppercase boldface symbols (\eg~$\mathbf{W}$) denote matrices;
 
\section{Preliminary}
\subsection{Compressed sensing}
Compressed sensing concerns the problem of recovering a signal from a small set of measurements. In our case, we're interested in reconstructing the depth map $\d$ with full resolution from the sparse depth map $\d_s$ produced by LiDAR. To achieve this, certain prior knowledge of the signal is required. The most widely used prior assumption is that the signal can be reconstructed with a sparse linear combination of basis elements from an over-complete dictionary $\W$. This gives an optimization problem similar to sparse coding:
\begin{equation}
    \min_{\z} \Vert \M\W\z - \d_s \Vert + b\left\Vert\z\right\Vert_0,
    \label{eq:cs}
\end{equation}
where $\z$ is the code, $\W\z$ produces our predicted depth map, and $\M$ is a diagonal matrix with 0 and 1s on its diagonal. It's used to mask out the unmeasured portions of the signal, such that the reconstruction error is only applied to the pixels which have been measured.

The key question to apply CS in Eq.~\ref{eq:cs} is: 1) For high dimensional signals such as the depth map, how to design the dictionary such that it encourages uniqueness of the code while still being computationally feasible; 2) How to learn the dictionary to get best reconstruction accuracy. In Sec.~\ref{sec:dccs}, we are going to show that the dictionary can be factored into a structure equivalent to performing multi-layer convolution, and that we can unroll the optimization of  Eq.~\ref{eq:cs} into a network similar to a deep recurrent neural network. This allows us to learn the dictionary together with other hyper-parameters (\eg $b$) through end-to-end training.

\subsection{Deep Component Analysis}
\label{sec:compr-sens-deep}


Equation (\ref{eq:cs}) can be generalized to multi-layered sparse coding in which one seeks a very high level sparse representation $\z_{\ell}$ such that $\d = \W_1\W_2\ldots \W_{\ell-1}\z_{\ell}$ and each intermediate product $\z_i = \W_i\W_{i+1}\ldots \W_{\ell-1}\z_{\ell}$ is also sparse. This formulation makes using a large effective dictionary computationally tractable, and when the dictionaries have a convolutional structure it allows for increased receptive fields while keeping the number of parameters manageable. This is further generalized to Deep Component Analysis (DeepCA) by the recent work of Murdock \etal which replaces the $\ell_0$ loss with arbitrary sparsity-encouraging penalties. The DeepCA objective function is stated in \cite{murdock} as:
\begin{equation}
  \label{eq:2}
  \min_{\{\z_i\}} \sum_{i=1}^{\ell} \frac{1}{2} \left\Vert \z_{i-1} - \W_i\z_i\right\Vert_2^2 + \Phi_i(\z_i),
\end{equation}
where the $\Phi_j$ are sparsity encouraging regularizers. Previous work has shown that the specific choice of $\Phi(\x) = I(\x > 0) + b\left\Vert\x\right\Vert_1$ yields optimization algorithms very similar to a feed-forward neural network with Relu activation functions. By using the ADMM algorithm to solve equation (\ref{eq:2}), Murdock \etal create Alternating Direction Neural Networks, a generalization of feed forward neural networks which internally solve optimization problems with the form of (\ref{eq:2}). Alternating Direction Neural Networks (ADNNs) perform the optimization in a fully differentiable manner and cast the activation functions of each layer as the proximal operators of penalty function $\Phi_i$ of that layer. This allows for learning the dictionaries $W_i$ and parameters $b$ through gradient descent and back propagation with respect to an arbitrary loss function on the sparse codes. To mirror neural networks, Murdock \etal  apply various loss functions to the highest level of codes, which take the place of the output layer in traditional NNs. In the following sections we will show how ADNNs can be adapted to the depth completion problem within the framework of compressed sensing.\\

\section{Deep Convolutional Compressed Sensing}
\label{sec:dccs}


\subsection{Inference}
\label{sec:deep-compr-sens}

Directly applying compressed sensing to the DeepCA objective gives
\begin{equation}
  \min_{\{\z_i\}} \frac{1}{2} \left\Vert \d_s - \M\W_1\z_1 \right\Vert_2^2  +  \sum_{i=2}^{\ell} \frac{1}{2} \left\Vert \z_{i-1} - \W_i\z_i\right\Vert_2^2 + \sum_{i=1}^{\ell}\Phi_i(\z_i), 
  \label{eq:dccs_no_mask}
\end{equation}
where $\d_s$ is the input sparse depth map. However, if we take the $\W_i$ to have a convolutional structure then an element $\z_1$ will not be recovered if its spatial support contains no valid depth samples. Thus, extracting the higher level codes is itself a missing data problem and can be written the same way. This gives the full Deep Convolutional Compressed Sensing objective:
\begin{equation}
  \label{eq:dccs}
  \min_{\{\z_i \vert i>0\}} \sum_{i=1}^{\ell} \frac{1}{2} \left\Vert \M_{i-1}\z_{i-1} - \M_{i-1}\W_i\z_i\right\Vert_2^2 + \Phi_i(\z_i).
\end{equation}
Here, to simplify notation, we merge the depth reconstruction cost (left term in Eq.~\ref{eq:dccs_no_mask}) and the reconstruction cost of the codes together, with $\z_0 = \d_s$ and $\M_0$ denotes the mask $\M$ used in (\ref{eq:dccs_no_mask}) . Each $\M_i$ is a mask encoding which elements of $\z_i$ had any valid inputs in their spatial support. In practice computing $\M_i$ is done with a maxpooling operation with the same stride and kernel size as the convolution represented by $\W_{i+1}^T$.\\

We solve (\ref{eq:dccs}) using the ADMM algorithm, which introduces auxiliary variables $\y_i$ that we constrain to be equal to the codes $\z_i$ as below:
\begin{equation}
\begin{aligned}
\min_{\{\y_i, \z_i \vert i>0\}} & \sum_{i=1}^{\ell} \frac{1}{2} \left\Vert \M_{i-1}\y_{i-1} - \M_{i-1}\W_i\z_i\right\Vert_2^2 + \Phi_i(\y_i)\\
\text{s.t.~~} & \z_i = \y_i.
\end{aligned}
\label{eq:admm_obj}
\end{equation}
Here, we again refer the input sparse depth $\d_s$ as $\y_0$. With this, the augmented Lagrangian of (\ref{eq:admm_obj}) with dual variables $\bm{\lambda}$ and a quadratic penalty weight $\rho$ is:
\begin{equation}
  L_\rho(\z, \y, \bm{\lambda}) = \sum_{i=1}^{\ell} \frac{1}{2} \left\Vert \M_{i-1}\y_{i-1} - \M_{i-1}\W_i\z_i\right\Vert_2^2 + \Phi_i(\y_i) + \bm{\lambda}^T_i(\z_i - \y_i) + \frac{\rho}{2}\left\Vert\z_i - \y_i\right\Vert_2^2.
\end{equation}
The ADMM algorithm then minimizes $L_\rho$ over each variable in turn, while keeping all others fixed. Following Murdock \etal we will incrementally update each layer instead of first solving for all $\z_i$ followed by all $\y_i$. They show this order leads to faster convergence. The ADMM updates for each variable are as follows:
\begin{enumerate}
\item
 At each iteration $t+1$, $\z_i$ is first updated by minimizing $L_\rho$ with the associated auxiliary variable $\y_i$ from the previous iteration, and $\z_{i-1}$ from the current iteration fixed:
  \begin{equation}
    \begin{aligned}
      \z^{[t+1]}_i & = \argmin_{\z_i} L_\rho(\z_i, \y^{[t+1]}_{i-1}, \y^{[t]}_i, \bm{\lambda}_i^{[t]})\\
      &= (\W_i^T\M_{i-1}^T\M_{i-1}\W_i + \rho \I)^{-1}(\W_{i}^T\M_{i-1}^T\M_{i-1}\W\y^{[t+1]}_{i-1} + \rho \y^{[t]}_{i} - \bm{\lambda}^{[t]}).
    \end{aligned}
    \label{eq:wupdate1}  
  \end{equation}

  This gives a fully differentiable update of $\z_i$ but the matrix inversion is computationally expensive, especially since $W_i$ is in practice very large. To deal with this problem we make the approximation that $\W_i$ is a Parseval tight frame~\cite{murdock}, that is we assume $\W_i\W_i^T = \I$. In addition to being common practice in autoencoders with tied weights, this assumption is also made by Murdock \etal and has previously been explicitly enforced in deep neural networks~\cite{moustapha}. We can then use the binomial matrix identity to rewrite the $\z_i$ update as:
  \begin{equation}
    \begin{aligned}
      \z^{[t+1]} &= \tilde{\y}_i^{[t]} + \frac{1}{1 + \rho}\W^T_i\M_{i-1}^T(\M_{i-1}\y^{[t+1]}_{i-1} - \M_{i-1}\W_i\tilde{\y}^{[t]}_i),
    \end{aligned}\label{eq:zupdate}
  \end{equation}
  where $\tilde{\y}_i^{[t]} \triangleq \y_i^{[t]} - \frac{1}{\rho}$.
\item
 Similarly, the update rule for the auxiliary variables $\y_i$ is:
  \begin{equation}
    \begin{aligned}
      \y_i^{[t+1]} & = \argmin_{\y_i} L_\rho(\z^{[t+1]}_i, \z^{[t]}_{i+1}, \y_i, \bm{\lambda}_i^{[t]})\\
      &= \phi_i\left(\frac{1}{1+\rho}\W_{i+1}\z^{[t]}_{i+1} + \frac{\rho}{1+\rho}(\z^{[t+1]}_i + \frac{\bm{\lambda}_i^{[t]}}{\rho})\right)\\
      \y_{\ell} &= \phi_i\left(\z_i^{[t]} + \frac{\bm{\lambda}_i^{[t]}}{\rho}\right).
    \end{aligned}\label{eq:yupdate}
  \end{equation}

  Here $\phi_i$ is the proximal operator associated with the penalty function $\Phi_i$. For appropriate choices of $\Phi_i$, $\phi_i$ is differentiable and can be computed efficiently. With this in mind, we choose $\Phi_i(\x) = I(\x > 0) + b\left\Vert\x\right\Vert_1$ so that $\phi_i(\x) = \text{ReLU}(\x - \frac{b}{\rho})$.
\item Finally the dual variable $\bm{\lambda}_i$ is updated by:
  \begin{equation}
    \label{eq:lupdate}
    \bm{\lambda}_i^{[t+1]} = \bm{\lambda}_i^{[t]} + \rho(\z_i^{[t+1]} - \y_i^{[t+1]}).
  \end{equation}
\end{enumerate}
\begin{algorithm}
  \SetKwInOut{Input}{Input}\SetKwInOut{Output}{Output}
  \Input{model parameters $\W_i,b_i$, iterations $T$, sparse depth $\y_0 = \d_s$, mask $\M$}
  \Output{sparse codes $\z^{[T]}_i$, predicted depth $\d_\text{pred}$}
  \For{$i \leftarrow 1$ \KwTo $\ell$}{
    $\z_i^{[0]} \leftarrow \W_i^T\M^T_{i-1}\y_{i-1}$\;
    $\y_i^{[0]} \leftarrow ReLU(\z_i^{[0]} - b/\rho)$\;
    $\bm{\lambda}_i^{[0]} \leftarrow 0$\;
  }
  \For{$t \leftarrow 1$ \KwTo $T$}{
    \For{$i \leftarrow 1$ \KwTo $\ell$}{
      Update $\z^{[t]}_i$ using equation (\ref{eq:zupdate})\;
      Update $\y^{[t]}_i$ using equation (\ref{eq:yupdate})\;
      Update $\bm{\lambda}^{[t]}_i$ using equation (\ref{eq:lupdate})\;
    }
   Predict $\d_\text{pred}$ using equation (\ref{eq:recon});
  }
  \caption{Deep Convolutional Compressed Sensing}
  \label{alg:dcsc}
\end{algorithm}

The full procedure is detailed in algorithm (\ref{alg:dcsc}).
As shown in above, all the operations used in the ADMM iteration are differentiable, and can be implemented with deep learning layers \eg convolution, convolution transpose, and ReLU. We unroll the ADMM iteration for a constant number of iterations $T$, and output our optimized code $\z_\ell$ for the last layer.
We can then extract our prediction of the depth map by applying the effective dictionary to the high level code $z_{\ell}$ as shown in equation (\ref{eq:recon}). This is different from the standard decoder portion of a deep autoencoder, where the nonlinear activations are applied in between each convolution. Our approach does not require this since the internal optimization of $z_{\ell}$ enforces equality constraints between layers, which is not the case for conventional autoencoders. We choose to reconstruct the depth from $z_{\ell}$ instead of a lower layer because its elements have the largest receptive field and therefore $z_{\ell}$ will have the fewest number of missing entries.
\begin{equation}
  \label{eq:recon}
  \d_\text{pred} = \W_1\W_2\ldots \W_{\ell}\z_{\ell}
\end{equation}


\subsection{Learning}
\label{sec:dictionary-learning}

With the ADMM update unrolled to $T$ iterations as described above, the entire inference procedure can be thought of as a single differentiable function:
\begin{equation}
    \d_\text{pred} = f_\text{DCCS}^{[T]}(\M, \d_s; \{\W_i, b_i\})
\end{equation}
Thus the dictionaries $\W_i$ and the bias term $b_i$ which are the parameters for $f_\text{DCCS}^{[T]}$ can be learned through  stochastic gradient descent over a suitable loss function. Using the standard sum of squared loss error, dictionary learning is formed as minimizing the depth reconstruction error $\L_\text{reconstruct}$:
\begin{equation}
  \label{eq:9}
  \min_{\{\W_i, b_i\}} \sum_{n=1}^{N}\left\Vert \d_\text{gt}^{(n)} - \M'^{(n)}  \d_\text{pred}^{(n)})\right\Vert
\end{equation}
Where $\d_\text{gt}^{(n)}$ is the ground truth depth map of the $n_\text{th}$ training example. We allow the ground truth depth map to have missing value by using mask $\M'^{(n)}$ to segment out the invalid pixels in the ground truth depth map. 

In practice we found that due to the sparsity of the training data, the depth maps our method predicted were rather noisy. To fix this issue we included the well known anisotropic total variation loss (TV-L1) when training to encourage smoothness of the predicted depth map. Note that this change has no significant impact on the quantitative error metrics, but produces more visually pleasing outputs. The total loss is then given by summation of the depth reconstruction loss and the TV-L1 smoothness loss, with hyper-parameter $\alpha$ to control the weighting for the smoothness penalty:

\begin{equation}
  \label{eq:9}
  \L = \L_{reconstruct} + \alpha \L_\text{TV-L1}.
\end{equation}

We empirically determined that $\alpha = 0.1$ produces the best results.
\section{Experiments}

\subsection{Implementation Details}
\label{sec:impl-deta}

We implemented three variants of algorithm (\ref{alg:dcsc}) for the cases $\ell = 1,2,3$. For the single layer case we let $\W^T_1$ be a 11x11 convolution with striding of 2 and 8 filters. For $\ell = 2$ we let $\W^T_{1}$ be an 11x11 convolution with 8 filters and $W^T_{1}$ be a 7x7 convolution with 16 filters. Finally for the $\ell = 3$ case: $\W^T_{1}$ is an 11x11 convolution with 8 filters, $\W^T_{2}$ is a 5x5 convolution with 16 filters, and $\W^T_{3}$ is a 3x3 convolution with 32 filters. For both $\ell = 2$ and $\ell = 3$, all convolutions have striding of 2. For the single layer case we learned the dictionaries with the number of iterations set to 5 and then at test time increased the number of iterations to 20. For the two and three layer cases the number of iterations was fixed at train and test time to 10 except in section \ref{sec:effect-iter-optim} where the number of test and training iterations is varied. All training was done with the ADAM optimizer with the standard parameters: learning rate = 0.001,$\beta_1 = 0.9$, $\beta_2$ =0.999, $\epsilon = 10^{-8}$.

\subsubsection{Error Metrics}
\label{sec:error-metrics}

For evaluation on the KITTI benchmark we use the conventional error metrics~\cite{uhrig,Geiger2012CVPR}, \eg root mean square error (RMSE), mean absolute error (MAE), mean absolute relative error (MRE). We also use the percentage of inliers metric, $\delta_i$ which counts the percent of predictions whose relative error is within a threshold raised to the power $i$. Here, we use smaller thresholds ($1.01^i$) compared to the more widely used ones ($1.5^i$) in oder to compare differences in performance under tighter metrics.

\subsection{KITTI Depth Completion Benchmark}
\label{sec:kitti-depth-compl}
\begin{table}
\centering
\begin{tabular}{r|cccccc}
  & RMSE (m) & MAE (m) & MRE & $\delta_1<1.01$ & $\delta_2<1.01^2$ & $\delta_3<1.01^3$\\\hline
  Bilateral NN\cite{Jampani2016CVPR} & 4.19 & 1.09 & - & - & - & -\\
  SGDU\cite{schneider2016semantically} & 2.5 & 0.72 & - & - & - & -\\
  Fast Bilateral Solver\cite{barron2016fast} & 1.98 & 0.65 & - & - & - & -\\
  TGVL\cite{Ferstl2013} & 4.85 & 0.59 & - & - & - & -\\\hline
  Closest Depth Pooling & 2.77 & 0.94 & - & - & - & -\\
  Nadaraya Watson\cite{Nadaraya,Watson} & 2.99 & 0.74 & - & - & - & -\\
  ConvNet & 2.97 & 0.78 & - & - & - & -\\
  ConvNet + mask & 2.24 & 0.79 & - & - & - & -\\
  SparseConvNet\cite{uhrig} & 1.82 & 0.58 & 0.035 & 0.33 & 0.65 & 0.82\\
  Ma \& Karaman\cite{sparsetodense}& 1.68 & 0.70 & 0.039 & 0.21 & 0.41 & 0.59\\\hline
  Ours 1 Layer & 2.77 & 0.83 & 0.054 & 0.3 & 0.47 & 0.59\\
  Ours 2 Layers & 1.45 & 0.47 & 0.028 & 0.41 & 0.68 & 0.8\\
  Ours 3 Layers & \textbf{1.35} & \textbf{0.43} & \textbf{0.024} & \textbf{0.48} & \textbf{0.73} & \textbf{0.83}\\
  \hline
 \end{tabular}
 \caption{Validation error of various methods on the KITTI Depth Completion benchmark. All results except for SparseConvNet and Ma's are taken as reported from ~\cite{uhrig}. Our method outperforms all previous state-of-the-art depth only completion methods (Middle) as well as those that use RGB images for guidance (Top).}
  \label{tab:kitti}
\end{table}

We evaluate our method on the new KITTI Depth Completion Benchmark~\cite{uhrig} instead of directly comparing against the LiDAR measurements from the raw KITTI dataset. The raw LiDAR points given in KITTI are corrupted by noise, motion of the vehicle during sampling, image rectification artifacts, and accounts to only 4\% of the total number of pixels in the image. Thus it's not ideal for evaluating depth completion systems. Instead, the benchmark proposed in~\cite{uhrig} resolved these issues by accumulating LiDAR measurements from nearby frames in the video sequences, and automatically removing accumulated LiDAR points that deviate too far from the points reconstructed by semi-global matching. This provides quality ground truth and effectively simulates the main application of interest: recovering dense depth from a single LiDAR sweep.



In Table~\ref{tab:kitti}, we form a close comparison against the very deep Sparse-to-Dense network (Ma \& Karaman~\cite{sparsetodense}) and the Sparsity Invariant CNN (SparseConvNet~\cite{uhrig}) which are the current sate-of-the-art deep learning-based method.


The Sparse-to-Dense network uses a similar deep network architecture as those used for single shot depth prediction -- with Resnet-18 as the encoder and up-projection blocks for the decoder. 
While the Sparse-to-Dense network is able to achieve good RMSE, it falls behind the SparseConvNet on MAE. We believe that this is because the deeper network can better estimate the average depth of a region but is unable to predict fine detail, leading to a higher MAE. By comparison, our method is able to both estimate the correct average depth and reconstruct fine detail due to its ability to directly optimize the prediction with respect to the input. Most notably our method outperforms all of the existing methods by a wide margin, including those that use RGB images and those that use orders of magnitude more parameters than our method.

\subsubsection{Varying Sparsity Levels}
\label{sec:vary-spars-levels}
\begin{figure*}
  \centering
  \begin{minipage}{0.45\textwidth}
    \centering
    \includegraphics[width=\linewidth]{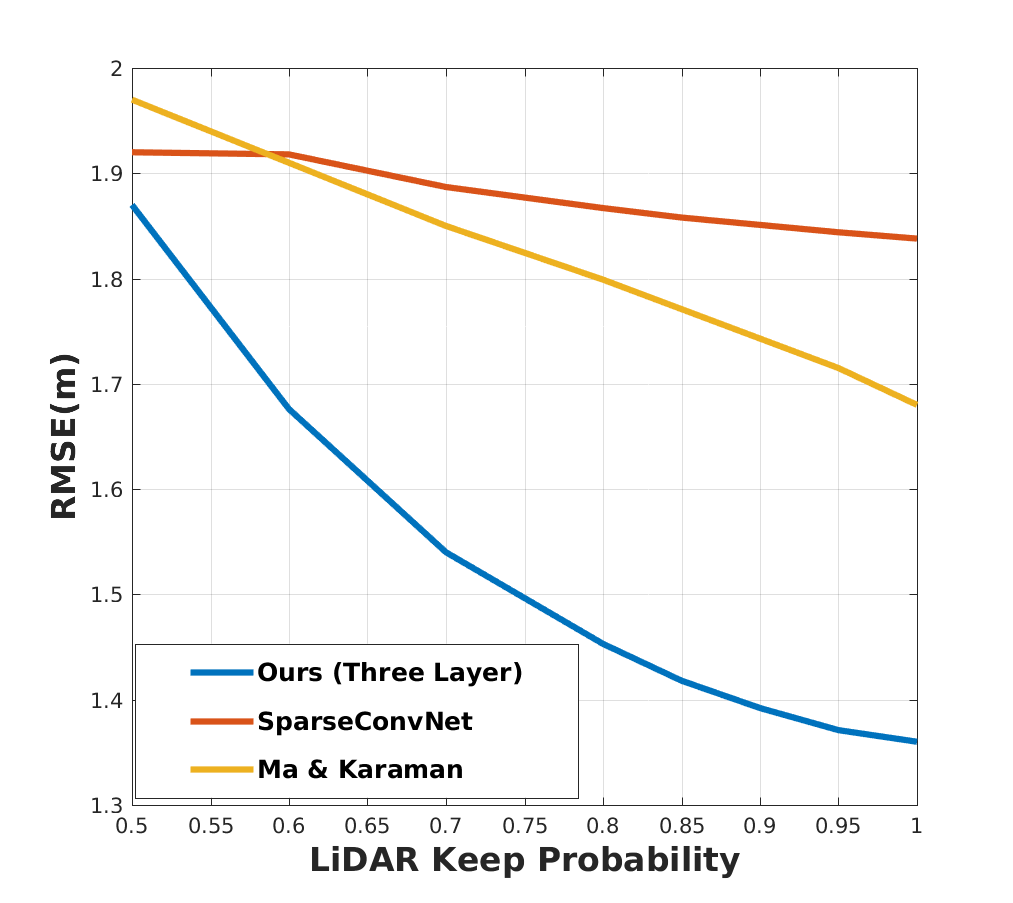}
    \captionof{figure}{Results on the KITTI benchmark for varying levels of input sparsity. The keep probability represents the probability that any particular LiDAR sample is retained. We demonstrate robustness to reasonable changes in input sparsity, outperforming both baselines up to a 50\% reduction in the number of input points.}
    \label{fig:sparsity}
  \end{minipage}\hspace{1cm}
  \begin{minipage}{0.45\textwidth}
    \centering
    \includegraphics[width=\linewidth]{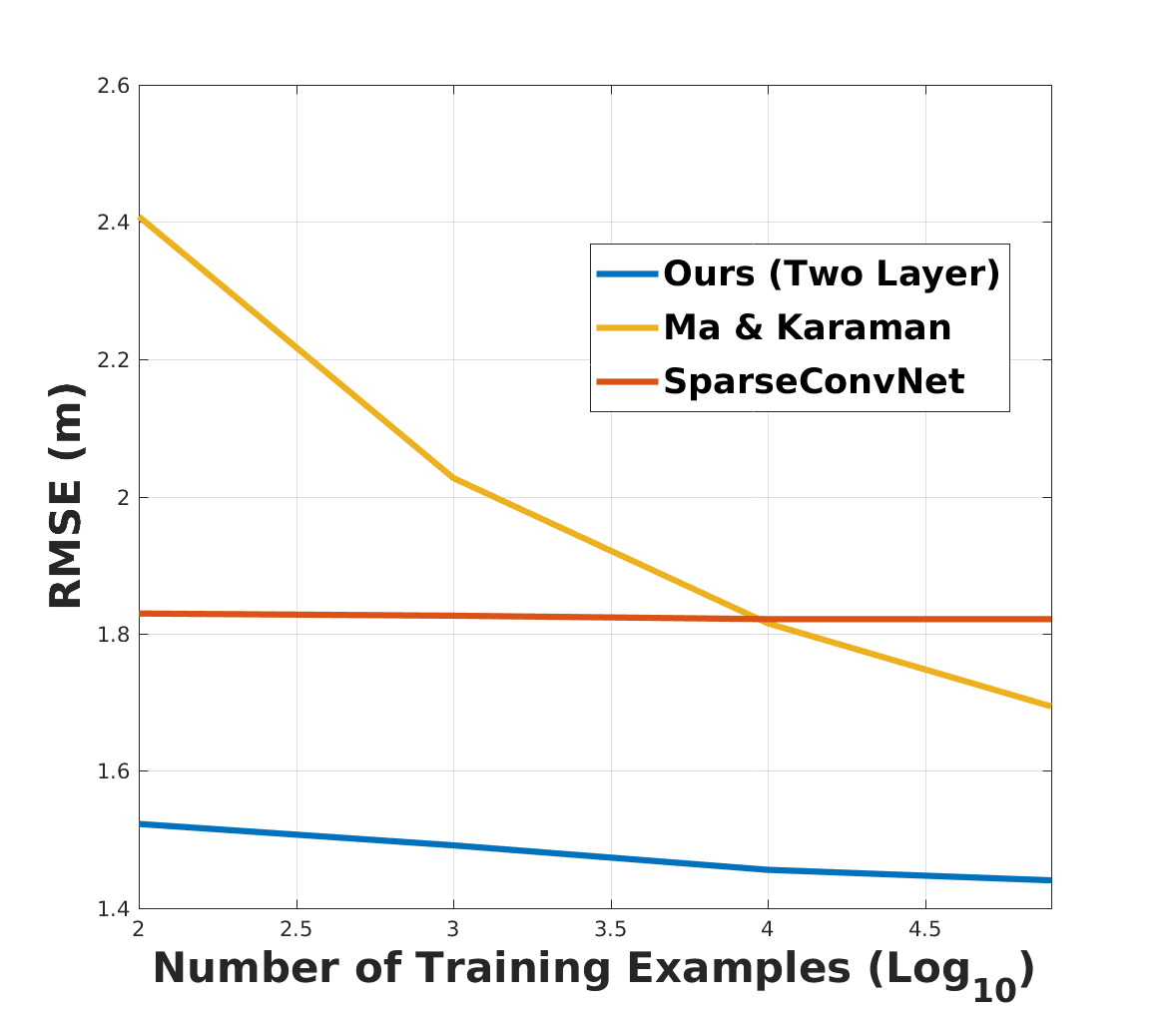}
    \caption{Results of selected methods on the KITTI benchmarks for varying training set sizes. Our method performs well with training sizes ranging from 100-86k but still benefits from larger training sizes.}
    \label{fig:trainsize}
  \end{minipage}
\end{figure*}

Uhrig \etal show that their Sparsity Invariant CNNs are very robust to a mismatch between the training and testing levels of sparsity. While we do not see a practical use for disparities as large as those tested in ~\cite{uhrig}, we do believe that depth completion systems should perform well under reasonable sparsity changes. To this end we adjusted the level of input sparsity in the KITTI benchmark by dropping input samples with probability $p$, for various values of $p$. The results of this experiment are shown in Figure (\ref{fig:sparsity}). While it is clear that our method does not achieve the level of sparsity invariance of the SparseConvNet, it still outperforms both baseline results even when the only 50\% of the input samples are kept.
\subsection{Effect of Amount of Training Data}
\label{sec:effect-training-data}

Modern deep learning models typically have tens of thousands to millions of parameters and therefore require enormous training sets to achieve good performance. This is in fact the motivation for the KITTI depth completion dataset, since previous benchmarks did not have enough data to train deep networks. In this section we investigate the dependence on the amount of training data on the performance of our method in comparison with a standard deep network and the sparsity invariant variety.

Figure (\ref{fig:trainsize}) shows the results of evaluating these models on the 1k manually selected validation depth maps after training on varying subsets of the 86k training maps. Our method outperforms both baselines for all training sizes. As expected Ma \& Karaman's method fails to generalize well when trained on a small dataset since the model has ~3.4M parameters but performs well once trained on the full dataset. It is interesting to observe that the method of Uhrig \etal does not gain any performance from training on more data. As a result it is ultimately out performed by the deep network which does not take sparsity into account. Our method is able to perform comparably to the sparsity invariant network with only 100 training examples but does increase in performance when given more data, validating the need for learning layered sparse coding dictionaries from large training sets. 
\subsection{Effect of Iterative Optimization}
\label{sec:effect-iter-optim}
\begin{figure}
  \centering
  \includegraphics[width=0.7\textwidth]{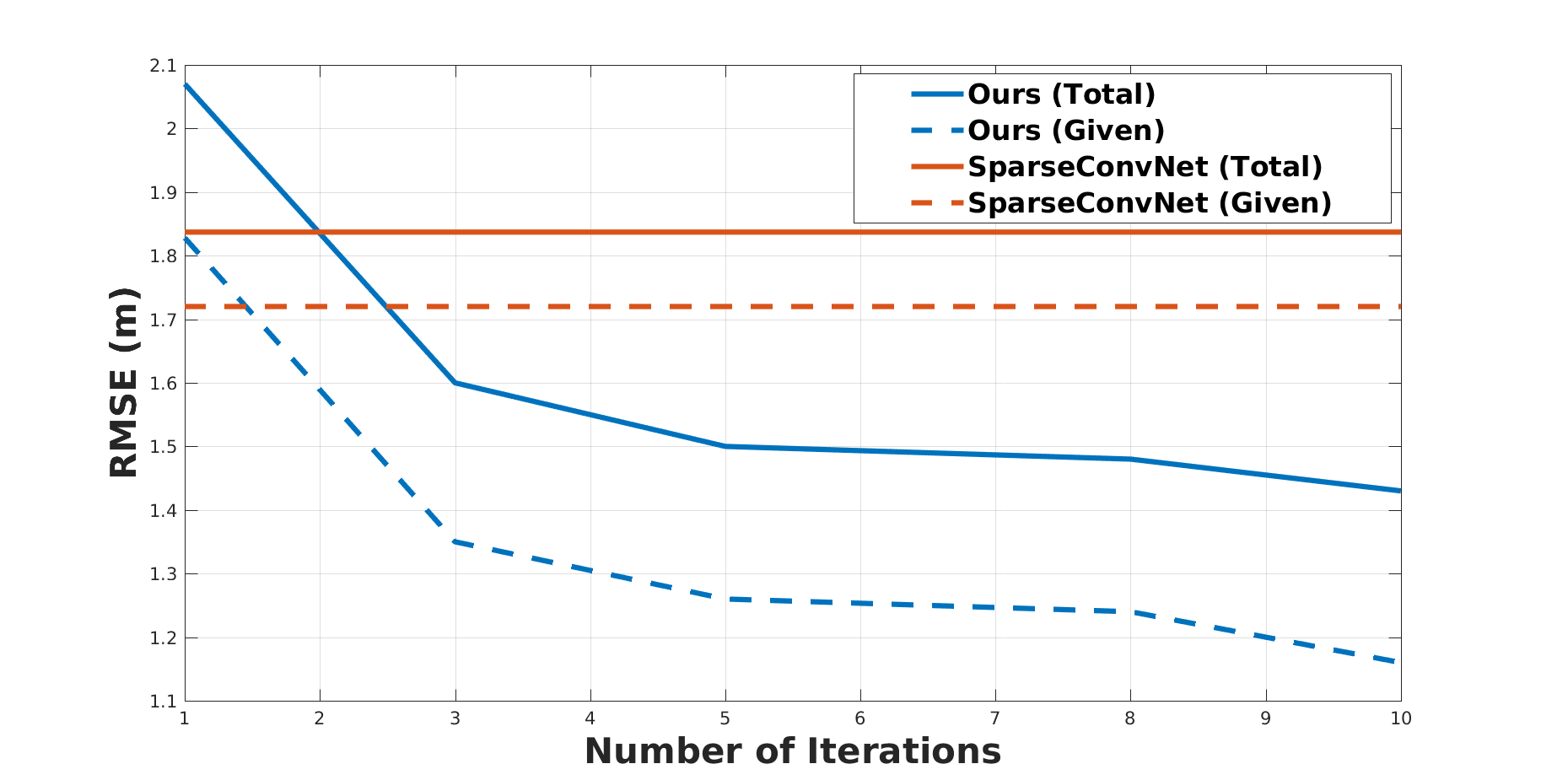}
  \caption{Results on the depth completion benchmark for different numbers of ADMM iterations. The total error is shown in blue while the red line shows the error on just those points given as input. The dotted lines show the same metrics but for the SparseConvNet of Uhrig \etal\cite{uhrig}.}
  \label{fig:iterplot}
\end{figure}

In this section we demonstrate that the success of our approach comes from its ability to refine depth estimates over multiple iterations. Applying a feed forward neural network to this problem frames it as finding a mapping from sparse LiDAR points to true depth maps. This is a reasonable approach but it doesn't utilize all of the available information, specifically it doesn't encode the relationship that input samples are a subset of the output that has been corrupted by noise. In contrast, our approach of phrasing depth completion as a compressed sensing missing data problem directly expresses that relationship. By solving this problem in an iterative fashion our network that is able to find depth maps that are both consistent with the input constraints and have sparse representations.\\

The importance of iterative optimization is shown in Figure (\ref{fig:iterplot}) where we examine the performance of our method as a function of the number of ADMM iterations it uses. It is clear that with few iterations our network fails to enforce the constraints and performs comparably to the SparseConvNet.
This is also consistent with Murdock \etal's observation that a feed forward network resembles a single iteration of an ADNN. As we increase the number of iterations our method is able to better optimize its prediction and gains a substantial performance boost.


\section{Conclusion}
\label{sec:conclusion}

In this work we have proposed a novel deep recurrent autoencoder for depth completion. Our architecture builds on the work of Murdock \etal on Deep Component Analysis and further establishes the link between sparse coding and deep learning. We demonstrate that our model outperforms existing methods for depth completion, including those that leverage RGB information. We also show that the success of our method is fundamentally a product of the internal optimization it performs, and that due to its small number of parameters it is able to perform well even without a large training set.\\

\begin{figure}[t]
  \centering
  \fbox{\begin{minipage}{0.45\linewidth}
      \includegraphics[width=\linewidth]{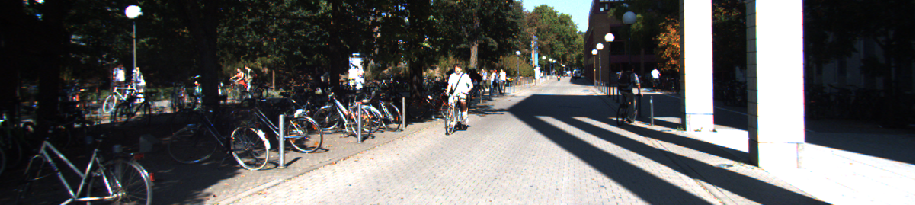}
      \includegraphics[width=\linewidth]{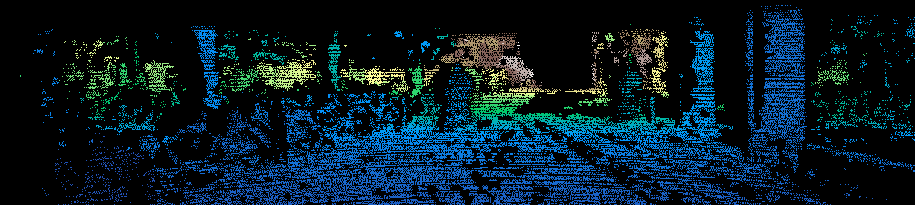}
      \includegraphics[width=\linewidth]{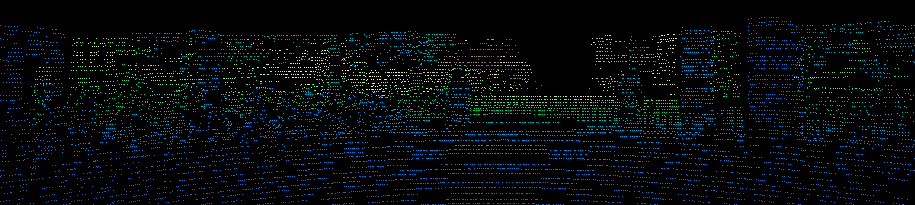}
      \includegraphics[width=\linewidth]{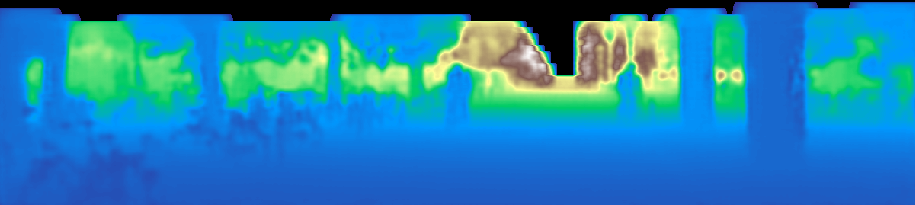}
    \end{minipage}}\hspace{1mm}
  \fbox{\begin{minipage}{0.45\linewidth}
      \includegraphics[width=\linewidth]{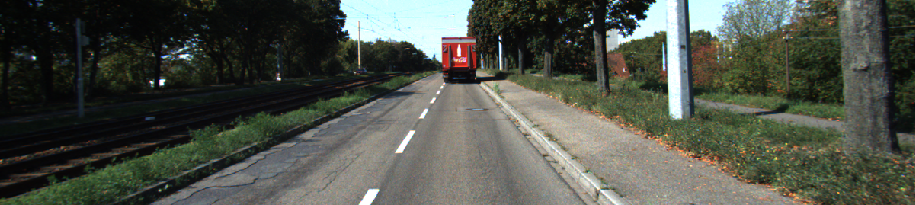}
      \includegraphics[width=\linewidth]{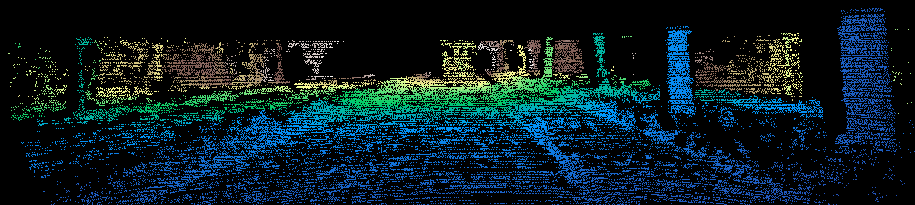}
      \includegraphics[width=\linewidth]{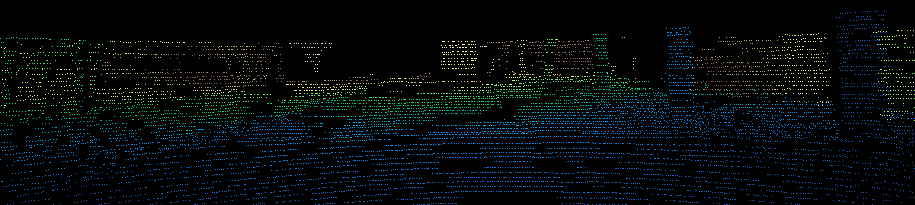}
      \includegraphics[width=\linewidth]{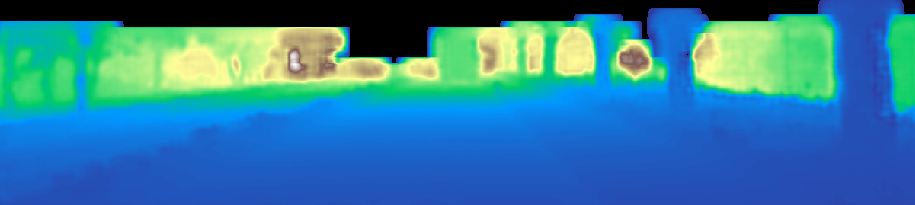}
    \end{minipage}}\\\vspace{1mm}
  \fbox{\begin{minipage}{0.45\linewidth}
      \includegraphics[width=\linewidth]{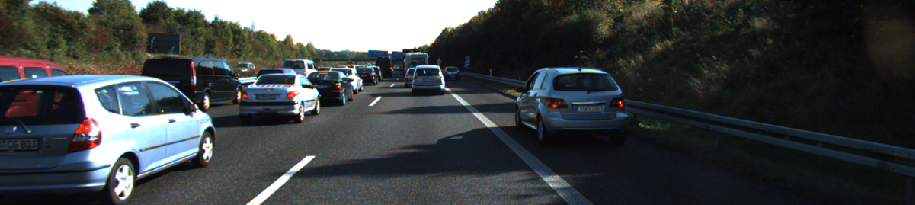}
      \includegraphics[width=\linewidth]{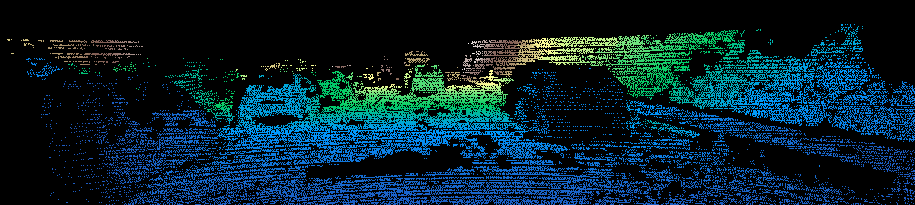}
      \includegraphics[width=\linewidth]{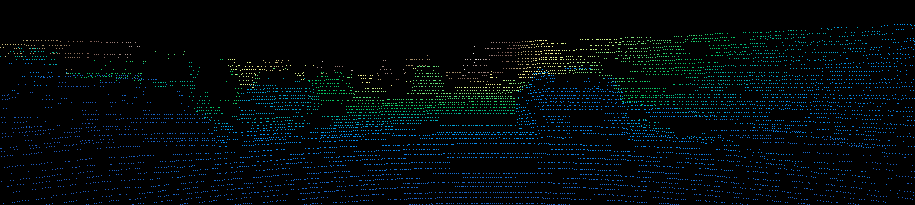}
      \includegraphics[width=\linewidth]{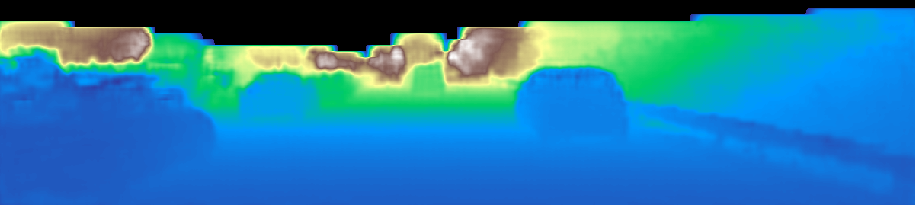}
    \end{minipage}}\hspace{1mm}
  \fbox{\begin{minipage}{0.45\linewidth}
      \includegraphics[width=\linewidth]{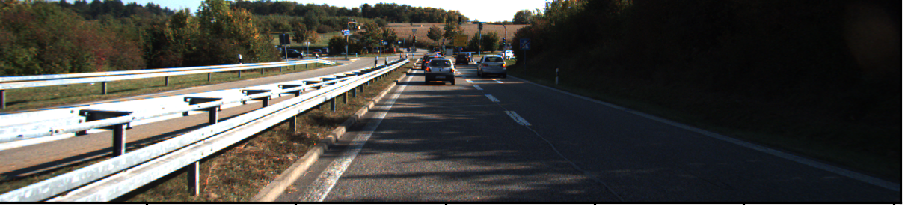}
      \includegraphics[width=\linewidth]{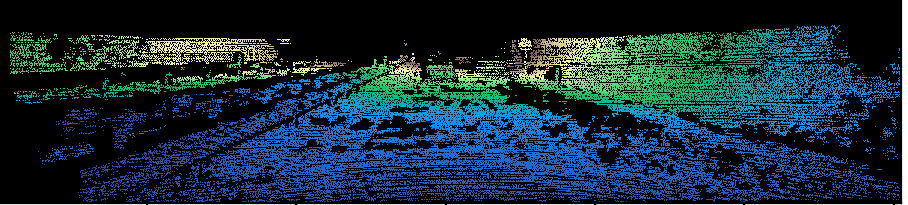}
      \includegraphics[width=\linewidth]{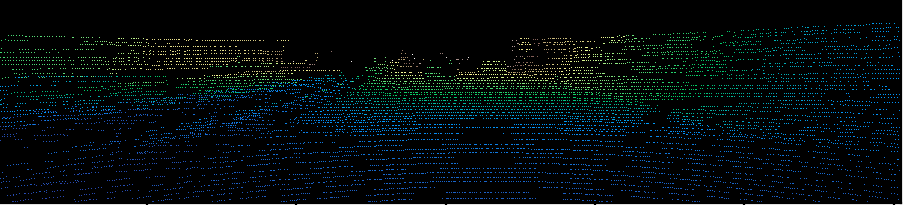}
      \includegraphics[width=\linewidth]{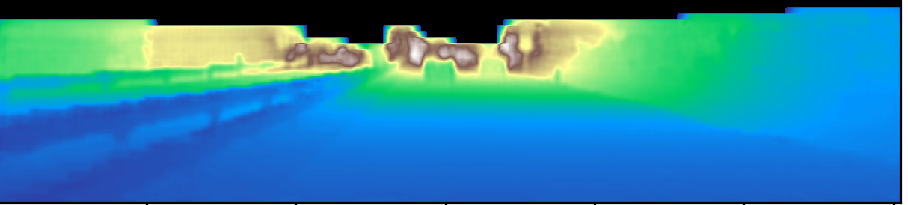}
    \end{minipage}}\\\vspace{1mm}
  \fbox{\begin{minipage}{0.45\linewidth}
      \includegraphics[width=\linewidth]{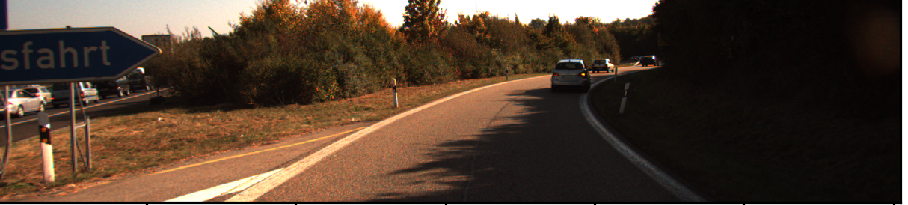}
      \includegraphics[width=\linewidth]{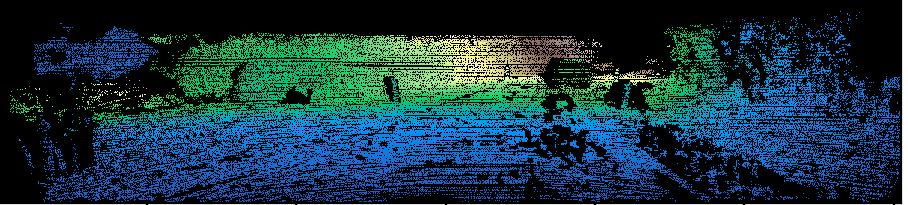}
      \includegraphics[width=\linewidth]{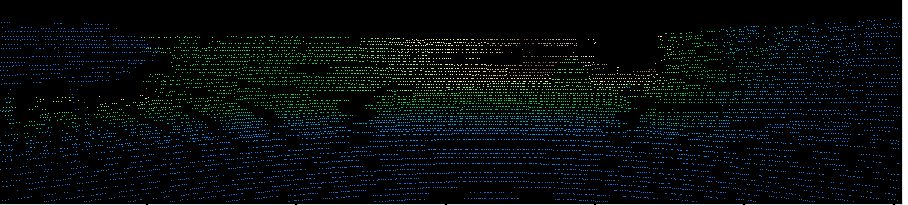}
      \includegraphics[width=\linewidth]{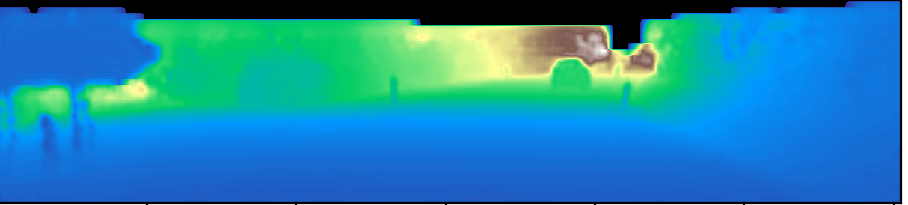}
    \end{minipage}}\hspace{1mm}
  \fbox{\begin{minipage}{0.45\linewidth}
      \includegraphics[width=\linewidth]{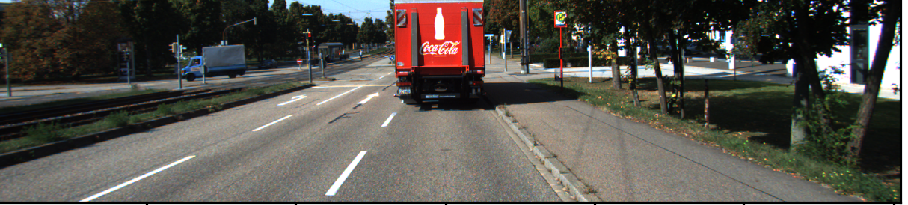}
      \includegraphics[width=\linewidth]{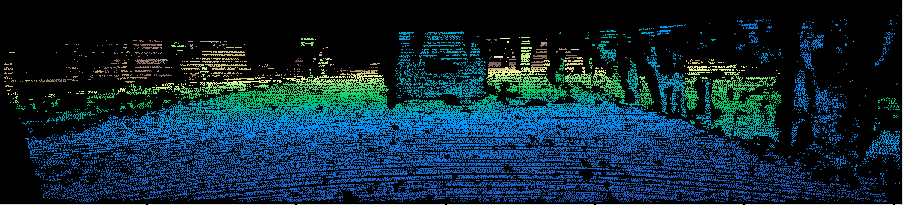}
      \includegraphics[width=\linewidth]{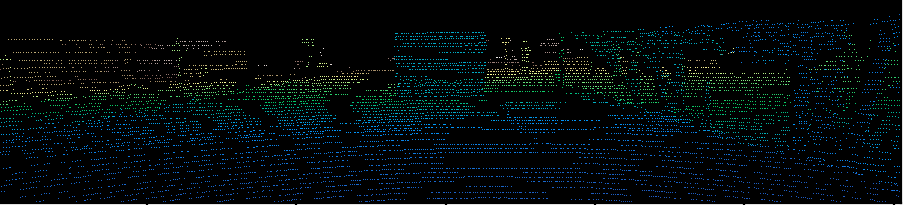}
      \includegraphics[width=\linewidth]{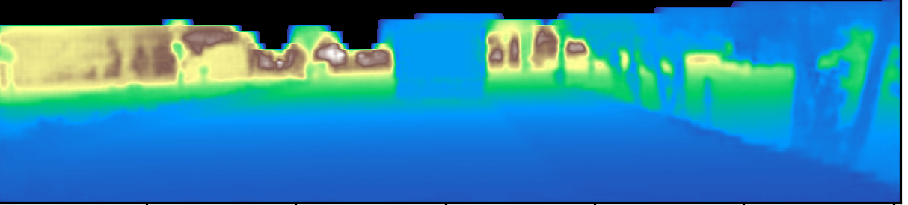}
    \end{minipage}}  
  \caption{Selected visual results form the KITTI benchmark. From top to bottom: RGB Image, Ground truth, input LiDAR points, Predicted depth.}
\end{figure}

\end{document}